\documentclass[10pt,twocolumn,letterpaper]{article}

\usepackage{cvpr}              

\usepackage[accsupp]{axessibility}

\usepackage{url}            
\usepackage{booktabs}       
\usepackage{amsfonts}       
\usepackage{nicefrac}       
\usepackage{microtype}      
\usepackage{xcolor}         

\usepackage{microtype}
\usepackage{graphicx}
\usepackage{booktabs} 

\usepackage{amsmath}
\usepackage{amssymb}
\usepackage{mathtools}
\usepackage{amsthm}

\usepackage{enumitem}
\usepackage{booktabs}
\usepackage{multirow}
\usepackage{multicol}
\usepackage{makecell}
\usepackage{array}
\usepackage{comment}
\usepackage{caption}

\usepackage{wrapfig}

\usepackage{bm}
\usepackage{tabularx}
\usepackage{colortbl}

\definecolor{brightpink}{rgb}{1.0, 0.0, 0.5}
\definecolor{ao(english)}{rgb}{0.0, 0.5, 0.0}
\definecolor{blue(ncs)}{rgb}{0.0, 0.53, 0.74}
\definecolor{babypink}{rgb}{0.96, 0.76, 0.76}
\definecolor{successColor}{rgb}{0.74, 0.83, 0.9}
\definecolor{blizzardblue}{rgb}{0.67, 0.9, 0.93}
\newcommand{\myPink}[1]{\textcolor{brightpink}{#1}}

\newcommand{\mybold}[1]{\textbf{#1}}

\newcommand{\SecModel}{\text{D4R}}
\newcommand{\AdaptSecModel}{\text{D4R-Align}}

\definecolor{cvprblue}{rgb}{0.21,0.49,0.74}
\usepackage[pagebackref,breaklinks,colorlinks,allcolors=cvprblue]{hyperref}

\def\paperID{2882} 
\def\confName{CVPR}
\def\confYear{2025}

\title{Mitigating the Human-Robot Domain Discrepancy in Visual Pre-training for Robotic Manipulation}

\author{Jiaming Zhou\textsuperscript{1},
Teli Ma\textsuperscript{1},
Kun-Yu Lin\textsuperscript{2},
Zifan Wang\textsuperscript{1},
Ronghe Qiu\textsuperscript{1},
Junwei Liang\textsuperscript{1,3}\footnotemark[2]
\\
\textsuperscript{1}{AI Thrust, The Hong Kong University of Science and Technology (Guangzhou)}\\ 
\textsuperscript{2}{Sun Yat-sen University,}
\textsuperscript{3}{The Hong Kong University of Science and Technology}\\
\tt\small jia\_ming\_zhou@outlook.com junweiliang@hkust-gz.edu.cn
}

\begin{document}

\twocolumn[{%
\renewcommand\twocolumn[1][]{#1}%
\maketitle

\begin{center}
    \centering
    \vspace{-0.6cm}
    \captionsetup{type=figure}
    \includegraphics[width=0.96\linewidth]{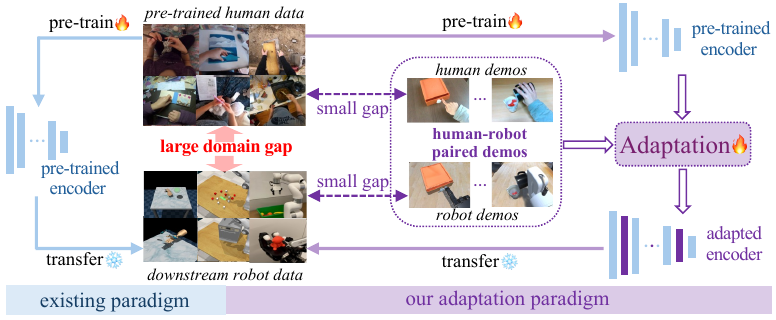}
    \caption{\textbf{Left}: The existing visual pre-training paradigm for robotic manipulation applied frozen models pre-trained on human data to downstream robot data, suffering from large domain gaps. \textbf{Right}: Our adaptation paradigm leverages the paired human-robot demonstrations as bridges to adapt the pre-trained models, thus mitigating the semantics gap between human and robot data. (Best viewed in color.)} 
    \label{fig:motivation}
\end{center}
}]

\footnotetext[2]{Corresponding author}

\begin{abstract}
Learning generalizable visual representations across different embodied environments is essential for effective robotic manipulation in real-world scenarios. 
However, the limited scale and diversity of robot demonstration data pose a significant challenge. 
Recent research has explored leveraging large-scale human activity data for pre-training, but the substantial morphological differences between humans and robots introduce a significant human-robot domain discrepancy, hindering the generalization of these models to downstream manipulation tasks.
To overcome this, we propose a novel adaptation paradigm that leverages readily available paired human-robot video data to bridge the domain gap. 
Our method employs a human-robot contrastive alignment loss to align the semantics of human and robot videos, adapting pre-trained models to the robot domain in a parameter-efficient manner.
Experiments on 20 simulated tasks across two different benchmarks and five real-world tasks demonstrate significant improvements.
These results span both single-task and language-conditioned multi-task settings, evaluated using two different pre-trained models.
Compared to existing pre-trained models, our adaptation method improves the average success rate by over $7\%$ across multiple tasks on both simulated benchmarks and real-world evaluations.
Project: \url{https://jiaming-zhou.github.io/projects/HumanRobotAlign}
\end{abstract}

\section{Introduction}
\label{sec:intro}

Robotic manipulation~\cite{vuong2023open, shridhar2022cliport, kalashnikov2018scalable, sontakke2024roboclip, chen2023polarnet, goyal2023rvt, chi2023diffusion, nair2022r3m} aims to enable robots to acquire a diverse range of skills applicable to real-world scenarios. 
However, this task is inherently challenging as it requires accurately modeling the sequential semantics of visual observations to predict the corresponding end-effector actions.
Due to the scarcity of robot demonstration data, various approaches~\cite{nair2022r3m, xiao2022masked, huo2023human, dasari2023unbiased, karamcheti2023language, ma2023liv, he2024large} have explored pre-training visual models on large-scale human activity datasets~\cite{shan2020understanding, grauman2022ego4d, goyal2017something, deng2009imagenet, damen2020epic}, such as Ego4D~\cite{grauman2022ego4d}. 
These pre-trained models are then used as frozen visual backbones to facilitate learning manipulation policies for downstream robotic tasks.
The central intuition~\cite{bahl2022human} underlying this paradigm is the dynamic similarity between human-object interactions and robot actions in similar tasks. 
This similarity facilitates the application of visual models, learned from human data, to downstream robotic manipulation tasks.

Nonetheless, the intrinsic morphological differences between humans and robots lead to a substantial domain discrepancy between human and robot data~\cite{ganin2016domain, ben2006analysis}.
As depicted in the left part of Figure~\ref{fig:motivation}, the existing paradigm cannot ensure an effective transfer of models pre-trained on extensive human data to downstream manipulation tasks.
Recently, two primary approaches have been proposed to solve this problem. 
The first~\cite{huo2023human, bahl2023affordances} focuses on introducing manipulation-oriented proxy tasks during pre-training on human data. 
For instance, Huo et al.~\cite{huo2023human} proposed training perceptual tasks (e.g., hand detection) on top of pre-trained models to adapt their representations for downstream robotic policies. 
However, this method lacks explicit access to robot data during training, making it difficult to directly mitigate the domain discrepancy. 
Moreover, defining manipulation-oriented tasks on human data remains a complex challenge.
The second approach~\cite{sharma2022lossless, lin2023spawnnet} integrates learnable modules into pre-trained models and fine-tunes the models on downstream robot data to learn more discriminative representation. 
However, this approach requires customizing the pre-trained models for each distinct environment, which reduces the versatility of the pre-trained models.
As a result, a dilemma arises of adapting human-data pre-trained representations to the robot domain while maintaining the model's versatility.

In this work, we propose a new adaptation paradigm, which mitigates the human-robot domain discrepancy while maintaining the versatility of pre-trained models for different downstream manipulation tasks, as depicted in the right part of Figure~\ref{fig:motivation}. 
To achieve this, we leverage a small amount of paired human-robot demonstration data as a bridge for adapting the pre-trained models. 
Such data, readily available in the community (e.g., the RH20T dataset~\cite{fang2023rh20t}), exhibit two essential characteristics.
First, each pair of human and robot demonstrations has well-aligned semantics. 
Second, although the used robot demonstrations are visually different from downstream robot data, the domain discrepancy between them is reduced due to their homogeneous robot structure.
Leveraging the similar dynamic semantics within the paired human-robot demonstrations, the key insight motivating our adaptation is that the semantics in robot demonstrations, as captured by the adapted models, should be consistent with the pre-trained semantics in the paired human demonstrations.
This paradigm effectively alleviates the discrepancy between pre-trains and downstream robot data while preserving the versatility of the adapted models, thereby eliminating the need for customization for each downstream environment.
Building on this new paradigm, we introduce Human-Robot Semantic Alignment (HR-Align), a simple yet effective method for adapting pre-trained models. 
HR-Align incorporates a parameter-efficient adapter module~\cite{yang2023aim, pan2022st, sharma2022lossless} into the models to facilitate adaptation.
To guide the adaptation towards reducing the domain discrepancy, our HR-Align method employs a human-robot contrastive alignment loss to ensure semantic consistency between the representations of paired human and robot data.

The contributions of this work are three-fold:
\begin{itemize}[itemsep=0pt, topsep=0pt, parsep=0pt, partopsep=0pt]

\item We highlight the human-robot domain discrepancy problem in visual pre-training for robotic manipulation, and provide a new adaptation paradigm that simultaneously alleviates the domain discrepancy and maintains the versatility of pre-trained models;

\item We propose a Human-Robot Semantic Alignment method, which adapts pre-trained models with parameter-efficient design and exploits a human-robot contrastive alignment loss for effectively mitigating the domain discrepancy;

\item We demonstrate the effectiveness of our method on 20 simulated tasks across two benchmarks and five real-world tasks. Compared to existing pre-trained models, our method achieves over a $7\%$ improvement in average success rate across both single-task and multi-task settings.

\end{itemize}

\section{Related Works}
\label{sec:related_works}

\noindent\textbf{Visual-based robotic manipulation.}
Driven by the rapid evolution of computer vision~\cite{krizhevsky2012imagenet, he2016deep, zheng2025diffuvolume, zheng2024selective, Zhou_2021_CVPR, he2022masked, feichtenhofer2022masked}, the developments of robotic manipulation~\cite{vuong2023open, shridhar2022cliport, kalashnikov2018scalable, sontakke2024roboclip, Xu_2024_CVPR, weigrasp, Wang_2024_CVPR, wu2024economic, chen2023polarnet, goyal2023rvt, chi2023diffusion, nair2022r3m} in recent years are noteworthy. Initial investigations~\cite{kalashnikov2018scalable, haarnoja2018composable, liu2021deep, james2020rlbench} developed visuo-motor policies focused on reinforcement learning~\cite{kaelbling1996reinforcement, mnih2013playing, schulman2017proximal}. This allowed robots to learn from their environments, although high sample complexity and difficult reward engineering were involved. More recent studies~\cite{chen2023polarnet, goyal2023rvt, chi2023diffusion, he2024large, james2022q, shridhar2023perceiver} have geared towards learning policies by imitating actions in expert demonstrations, which has yielded promising results.
In addition, the robotic manipulation community has been further propelled by the introduction of new benchmarks. Transitioning from benchmarks with limited scales and diversity~\cite{yu2020meta, rajeswaran2017learning, gupta2019relay} to larger scale benchmarks~\cite{james2020rlbench, mees2022calvin, vuong2023open, mu2021maniskill} with the increased object, scene, and task variations (e.g., RLBench~\cite{james2020rlbench} and CALVIN~\cite{mees2022calvin}). 

\noindent\textbf{Robotic visual pre-training.} 
To efficiently teach robots manipulation skills in real-life scenarios, it is crucial to pre-train models on large-scale data. This ensures that the generalizable visual representation across different environments is effectively learned. Drawing inspiration from the representation learning~\cite{he2022masked, He_2020_CVPR, wang2023videomae} in computer vision, there have been several endeavors~\cite{brohan2022rt, brohan2023rt, vuong2023open, zhen20243d, li2023vision, yang2024pushing} to accomplish this goal in robotic manipulation.
The RT-series~\cite{brohan2022rt, brohan2023rt, vuong2023open}, for instance, developed vision-language-action models using large-scale data that encompassed out-of-domain data from the internet and self-gathered robotic demonstrations. 
RoboFlamingo~\cite{li2023vision} learned to generate robot actions by fine-tuning pre-existing vision-language foundation models on robot demonstrations. However, the collection of large-scale robot demonstration data proved costly, and unresolved issues remain regarding the different data distributions between different embodiments~\cite{chen2024mirage, yang2024pushing, vuong2023open}.

In parallel, several studies~\cite{nair2022r3m, xiao2022masked, dasari2023unbiased, he2024large, radosavovic2023real, jing2023exploring, ma2022vip, karamcheti2023language, ma2023liv, majumdar2024we} have proposed methods to learn a generalizable representation from large-scale human datasets, such as Ego4D~\cite{grauman2022ego4d} and Kinetics~\cite{carreira2018short}. These methods are based on the assumption that the dynamics of human-object interactions in these datasets are analogous to those of robots performing tasks. For instance, the R3M~\cite{nair2022r3m} model was pre-trained on the Ego4D dataset using time-contrastive learning and video-language alignment techniques. Similarly, the MVP~\cite{xiao2022masked} model employed masked modeling to learn visual dynamics, and the data4robotics~\cite{dasari2023unbiased} performed a dataset-centric analysis of robotic pre-training, contributing valuable insights to the field.
Despite these advances, significant challenges remain due to the inherent morphological differences between humans and robots. These differences lead to a human-robot domain discrepancy~\cite{huo2023human, lin2023spawnnet, sharma2022lossless, jain2024vid2robot, bahl2022human, xiong2023robotube}, which limits the generalizability of models trained on human data when applied to downstream robotic tasks.

\noindent\textbf{Human-robot domain transfer.} 
Many works~\cite{ma2023liv, schmeckpeper2020reinforcement, xiong2021learning, zakka2022xirl, bharadhwaj2024towards, bahl2022human, li2024ag2manip} have been dedicated to learning robotic manipulation skills by imitating human actions in human videos. For large-scale pre-training using human data, some works~\cite{huo2023human, sharma2022lossless, lin2023spawnnet} have noticed the human-robot domain gap problem and have developed two main strategies to address it.
The first is to design more effective proxy tasks to learn semantics that can be better transferred to downstream robotic manipulation when pre-training models on the human data. 
For example, Huo et al.~\cite{huo2023human} proposed human-oriented multi-task fine-tuning on top of pre-trained models to encode meaningful semantics for manipulation. However, the manipulation-oriented proxy tasks are difficult to define, and this implicit fine-tuning without accessing robot data is not effective in mitigating the human-robot domain discrepancy. 
The second strategy~\cite{sharma2022lossless, lin2023spawnnet} directly fine-tunes the pre-trained models on downstream manipulation data to bridge the semantic gap between robot data and pre-trained human data. 
For example, Sharma et al.~\cite{sharma2022lossless} proposed to insert lightweight adapters into pre-trained models, aiming to reduce the performance gap between frozen pre-trained representation and downstream full-parameter end-to-end fine-tuning. However, this strategy requires custom modifications to the pre-trained models for each specific environment, which will compromise the models' generalizability.

Beyond these two strategies, in this work, we propose a new approach with efficient human-robot semantic alignment, which adapts pre-trained representation to explicitly mitigate the human-robot domain discrepancy, while maintaining the versatility of the models across different robotic environments.

\section{Human-Robot Semantic Alignment Method}
\label{sec:method}

Considering the large domain discrepancy between the pre-trained human data and downstream robotic manipulation data in the existing pre-training paradigm, this work proposes a new adaptation paradigm. This paradigm leverages the semantic-aligned human-robot demonstration data to guide the adaptation of existing pre-trained models, effectively enhancing their generalization capabilities across various downstream tasks. Driven by this idea, we propose a simple yet effective Human-Robot Semantic Alignment method.

\subsection{Problem formulation}
Assuming we have a dataset $\mathcal{D}$ encompassing $N$ human-robot video pairs, i.e., $\mathcal{D}=\{H_i, R_i, L_i\}_{i=1}^N$, where $L_i$ is a sentence describing what task is being performed in the video pair, $H_i$ denotes the $i$-th human video demonstration, and $R_i$ denotes the $i$-th robot video demonstration that imitates the human actions in $H_i$ via teleoperation. Thus the dynamic semantics in each human-robot video pair are well-aligned. For learning the downstream manipulation policies, instead of using the robot's raw image observations $I$ as state input, existing visual pre-training methods train visual encoder $\mathcal{F}$ on human data, and use the feature representation $\mathcal{F}(I)$ of robot's observations as state input. In this work, to mitigate the human-robot domain discrepancy in the pre-trained representation of $\mathcal{F}$, we leverage the aforementioned paired human-robot video dataset $\mathcal{D}$ to adapt the pre-trained model $\mathcal{F}$. In this way, the adapted model $\mathcal{T}$ can extract more effective state representation for learning policies in downstream manipulation tasks.

\begin{figure*}[!t]
\centering
    \includegraphics[width=0.9\linewidth]{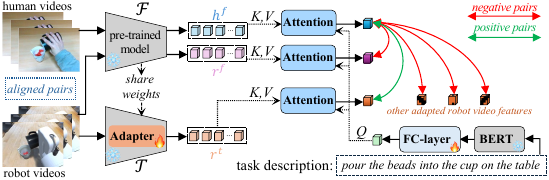}
    \caption{\textbf{An overview of the proposed Human-Robot Semantic Alignment method.} Given paired human-robot videos, the pre-trained models are efficiently adapted on the robot data to learn semantics aligned with those in human data.}
\label{fig:method}
\end{figure*}

\subsection{Overview}
Figure~\ref{fig:method} provides an overall diagram of our Human-Robot Semantic Alignment (HR-Align) method. 
Taking the paired human and robot videos as input, the HR-Align method utilizes a frozen pre-trained model to obtain the feature representations of human video and robot video, respectively. 
However, since the used backbone model is pre-trained on human data, the extracted feature semantics of human and robot videos suffer from domain discrepancy. 
For the robot video, our HR-Align injects learnable adapter modules into the pre-trained model, so that the pre-trained model can be adapted to effectively encode the dynamic semantics in the robot video.
Besides, for each stream, the corresponding task description of the videos will serve as the query to extract task-aware semantics from the aforementioned feature representations.
To guarantee the introduced adapter modules can be effectively trained for adapting the pre-trained model to the robot domain, the HR-Align leverages a contrastive alignment loss, which constraints the semantics in robot video encoded by the adapted model to align with the pre-trained feature semantics in the paired human video.

\subsection{HR-Align Learning}

\noindent\textbf{Frozen human and robot streams.} 
Given a model $\mathcal{F}$ pre-trained on human data, it takes as input a batch of $M$ paired human-robot videos and the corresponding task descriptions, i.e., $\{H_i, R_i, L_i\}_{i=1}^M$. 
For a cleaner presentation in the following, we take the $i$-th video pair as an example and omit the subscript $i$, i.e., the human video $H$, robot video $R$, and task description $L$. 
For both the human video $H$ and robot video $R$, we randomly sample $T$ frames and use the frozen pre-trained model $\mathcal{F}$ to extract their spatial-temporal features:
\begin{align}
h^f = \mathcal{F}(\Upsilon(H)), \quad \quad r^f = \mathcal{F}(\Upsilon(R)),
\end{align}
where $\Upsilon$ denotes the frame sampling operation, and $h^f, r^f \in\mathbb{R}^{\mathrm{\hat{T}\times \hat{H}\times \hat{W}\times \hat{C}}}$ are the frozen features extracted from the human video and robot video, respectively.
Since the backbone model $\mathcal{F}$ is well pre-trained on large-scale human data, the visual dynamic semantics in human video $H$ can be effectively captured. 
However, due to the significant human-robot domain gap, it is difficult to extract corresponding semantics from the paired robot video, when using the human-data trained model $\mathcal{F}$ that does not access any knowledge in the robot domain during pre-training.

\noindent\textbf{Adapted robot stream.} 
To this end, our HR-Align method aims to efficiently adapt the pre-trained model on the robot videos, such that the newly adapted pre-trained model $\mathcal{T}$ can alleviate the domain discrepancy.
To achieve this, inspired by existing parameter-efficient fine-tuning approaches~\cite{yang2023aim, pan2022st}, we form the adapted pre-trained model $\mathcal{T}$ by inserting learnable adapter modules (Eq.~\eqref{eq:adapter}) into the pre-trained model $\mathcal{F}$. 
Then we apply this model to the robot video as follows:
\begin{align}
r^t = \mathcal{T}(\Upsilon(R)) = \mathcal{F}_{\textit{Adapter}}(\Upsilon(R)),
\end{align}
where $r^t\in\mathbb{R}^{\mathrm{\hat{T}\times \hat{H}\times \hat{W}\times \hat{C}}}$ is the adapted robot video features. The adapter modules can be inserted into any position of the pre-trained model $\mathcal{F}$. Without loss of generality, given the output $r^{f, inter}\in\mathbb{R}^{\mathrm{\bar{T}\times \bar{H}\times \bar{W}\times \bar{C}}}$ of an intermediate layer of model $\mathcal{F}$, the adapter module can be applied as follows:
\begin{equation}
\begin{aligned}
\label{eq:adapter}
r^{t, next} &= \textit{Adapter}(r^{f, inter}) \\
&= r^{f, inter} + \text{Conv}_{up}(g(\text{Conv}_{down}(r^{f,inter}))),
\end{aligned}
\end{equation}
where the residual output $r^{t, next}$ will be processed by subsequent layers of the model $\mathcal{F}$. Following existing practice~\cite{yang2023aim, pan2022st}, our adapter module includes an activation function $g$, and two convolution layers (i.e., $\text{Conv}_{down}$ and $\text{Conv}_{up}$) used for feature down-projection and up-projection, respectively.

\noindent\textbf{Task-aware feature modeling.} 
With the extracted frozen human feature $h^f$, frozen robot feature $r^{f}$, and the adapted robot feature $r^{t}$, we incrementally contribute a feature enhancement module to obtain task-aware feature representations of human and robot videos, where the corresponding task description $L$ serves as the query.
The feature of task description query $l\in \mathbb{R}^{\hat{C}}$ is obtained using a frozen BERT model~\cite{devlin2018bert,sanh2019distilbert}, with a learnable fully connected layer followed to match the channel dimension of the video features. The process is formulated as $l = \textit{Linear}(\textit{Bert}(L))$.
Later, an attention-based aggregation is defined, where the task description feature $l$ serves as the query, and the video features in each stream serve as keys and values. 
Taking the reshaped video features $r^t\in\mathbb{R}^{\mathrm{(\hat{T}\cdot \hat{H}\cdot \hat{W})\times \hat{C}}}$ in the adapted robot stream as an example, the task-aware adapted robot feature $\bar{r}^{t}$ is defined as:
\begin{align}
\mathcal{A}^{r} &= softmax(r^t\cdot l)\in \mathbb{R}^{\hat{T}\cdot \hat{H}\cdot \hat{W}\times 1}, \\
\bar{r}^t &= (r^t)^T\cdot \mathcal{A}^{t}\in \mathbb{R}^{\hat{C}}.
\end{align}
For the frozen human and robot data stream, the task-aware frozen human feature $\bar{h}^f\in \mathbb{R}^{\hat{C}}$ and task-aware frozen robot feature $\bar{r}^f\in \mathbb{R}^{\hat{C}}$ can be obtained in the same way.
While not a core contribution, the ablation study in the appendix demonstrates that the feature enhancement effectively improves the model’s performance.

\noindent\textbf{Human-Robot contrastive alignment.}
To mitigate the human-robot domain discrepancy in human-data pre-training, our key idea is to finetune the pre-trained model with a parameter-efficient design, where the semantic-aligned human-robot video pairs serve as bridges for the adaptation.
For each human-robot video pair, the human-data pre-trained model can effectively encode the visual semantics in the human video. 
To guarantee the consistent semantics in the paired robot video can be captured by the adapted model, we propose a human-robot contrastive alignment loss to modulate the adaptation of the pre-trained model.

For a batch of $M$ paired human-robot video data, i.e., $\{H_i, R_i, L_i\}_{i=1}^M$, their task-aware feature representations $\{\bar{h}_i^f, \bar{r}_i^f, \bar{r}_i^t\}_{i=1}^M$ can be obtained according to the above processes.
The proposed contrastive alignment loss stems from two principles. 
First, for the $i$-th video pair, regarding the well-captured human feature $\bar{h}_i^f$, we encourage the adapted robot feature $\bar{r}_i^t$ to encode more consistent semantics than the unadapted robot feature $\bar{r}_i^f$. 
Second, for each human feature $\bar{h}_i^f$, we encourage its semantics to be more consistent with the paired adapted robot feature $\bar{r}_i^t$, compared to all other unpaired adapted robot features $\{\bar{r}^t_j\}_{j\neq i}$. 
Similarly, for each adapted robot feature $\bar{r}_i^t$, it should be more consistent with the paired human feature $\bar{h}_i^f$ than all other unpaired human features $\{\bar{h}^f_j\}_{j\neq i}$. 
In this way, the adapted model effectively encodes the discriminative visual semantics in robot demonstrations.
Based on these, we formulate the human-robot contrastive alignment loss as follows:
\vspace{-0.2cm}
\begin{small}
\begin{equation}
\begin{aligned}
\label{eq:contrastive_loss}
\mathcal{L} &= \frac{1}{2M} \sum_{i=1}^{M}-\log\frac{\mathcal{S}(\bar{h}_{i}^{f},\bar{r}_{i}^{t})}{\mathcal{S}(\bar{h}_{i}^{f},\bar{r}_{i}^{t}) + \mathcal{S}(\bar{h}_{i}^{f},\bar{r}_{i}^{f}) + \sum_{j=1}^{M}\mathcal{S}(\bar{h}_{i}^{f},\bar{r}_{j\neq i}^{t})}\\
&+ \frac{1}{2M} \sum_{i=1}^{M}-\log\frac{\mathcal{S}(\bar{r}_{i}^{t},\bar{h}_{i}^{f})}{\mathcal{S}(\bar{r}_{i}^{t},\bar{h}_{i}^{f}) + \mathcal{S}(\bar{r}_{i}^{f},\bar{h}_{i}^{f}) + \sum_{j=1}^{M}\mathcal{S}(\bar{r}_{i}^{t},\bar{h}_{j\neq i}^{f})}.
\end{aligned}
\end{equation}
\end{small}
In Eq.~\eqref{eq:contrastive_loss}, $\mathcal{S}$ is a similarity function defined as $\mathcal{S}(x,y)=\exp(x^Ty/\tau)$, where $\tau$ is a temperature factor.
Under the constraint of the contrastive alignment loss, the learnable adapter modules enable the pre-trained model to capture robotic dynamics, thereby simultaneously mitigating the domain discrepancy and avoiding the need to customize models for each downstream robotic environment.

\section{Experiments}
\label{sec:exps}

\subsection{Experimental Setups}
The experimental validation includes two stages. First, given existing human-data pre-trained models, we adapt the models for mitigating the human-robot domain discrepancy using the proposed HR-Align method. Second, the pre-trained models after adaptation are used as frozen visual backbones for learning manipulation policies on downstream tasks.

\noindent\textbf{Pre-trained Model Adaptation.}
To comprehensively evaluate the efficacy of the proposed adaptation method, we adapt two human-data pre-trained models that are widely used in the robotic manipulation community. One is the R3M model~\cite{nair2022r3m} pre-trained on Ego4D~\cite{grauman2022ego4d} with vision-language contrastive pre-training. 
The other one comes from data4robotics~\cite{dasari2023unbiased}, which is pre-trained on Kinetics~\cite{carreira2017quo} with MoCo pre-training~\cite{he2020momentum} (We refer to it as D4R model).
For the paired human-robot video data, we instantiate it using the existing RH20T dataset~\cite{fang2023rh20t}. 
We select a subset from the RH20T dataset, which includes a total of $56$k human-robot video pairs.  
During adaptation, we randomly sample $5$ frames from each video (i.e., $T=5$). Following R3M~\cite{nair2022r3m}, we use the frozen DistilBert model~\cite{sanh2019distilbert}  to process the task descriptions. 
The Adam optimizer~\cite{kingma2014adam} with the learning rate of $1\times e^{-4}$ and batch size of $200$ is used during training. The temperature factor $\tau$ is set to $0.1$.
Without specific statements, we insert the adapter module after the last layer of the backbone network. 
The adaptation process takes about $8$k training steps with $4$ NVIDIA A6000 GPUs, requiring a small extra cost compared to human-data pre-training.

\noindent\textbf{Downstream Policy Learning.} 
To learn manipulation policies for different downstream tasks, the pre-trained model after adaptation is used to extract the state representations of the robot's visual observations. Then, action prediction models are built upon these frozen visual features to predict the next position of the robot's end-effector. 

\begin{figure*}[!ht]
  \centering
  \setlength{\abovecaptionskip}{0.1cm}
   \includegraphics[width=1.0\linewidth]{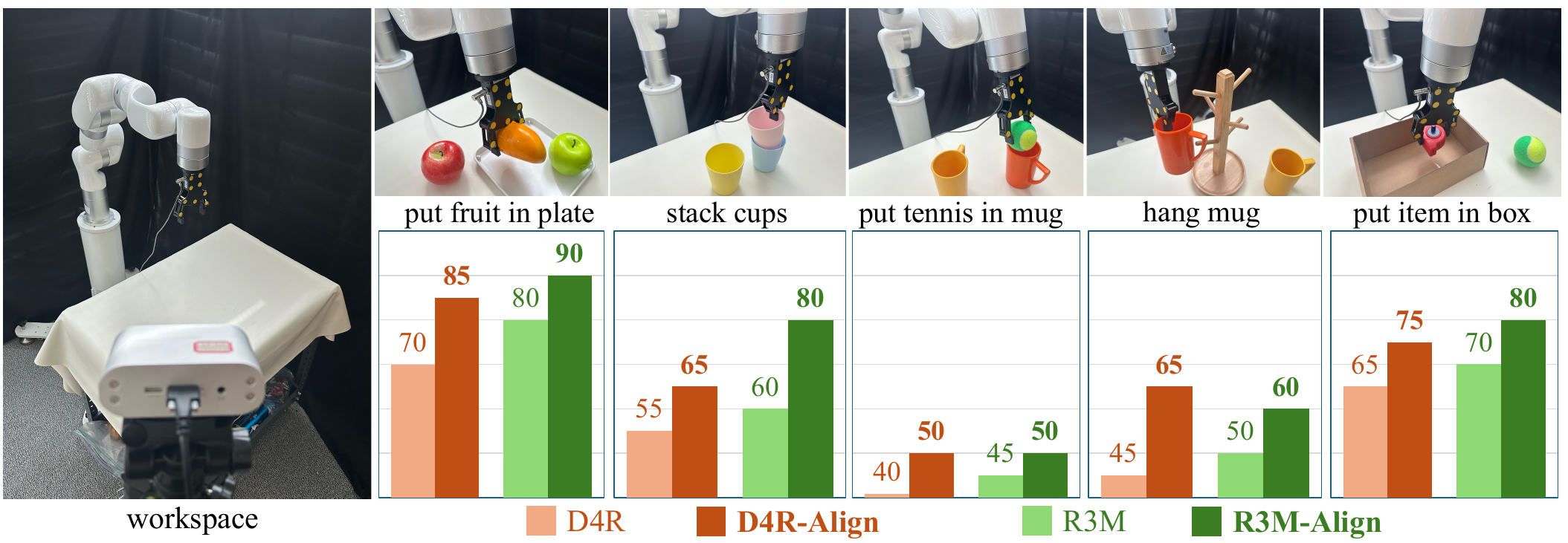}
   \caption{\textbf{Left}: our real-world experimental workspace. \textbf{Top right}: illustrations of five real-world tasks. \textbf{Bottom right}: experimental results on five tasks, where the pre-trained D4R and R3M, and our adapted D4R-Align, R3M-Align models are evaluated.}
\label{fig:real_world}
\vspace{-0.2cm}
\end{figure*}

\noindent\textit{\textbf{(i) Simulated tasks.}} 
For simulation environments, we evaluate a total of 20 robotic manipulation tasks across two different simulated benchmarks, including 2 single-tasks and 18 language-conditioned multi-tasks. 

For the single-task setting, following previous works~\cite{nair2022r3m, huo2023human}, two dexterous manipulation tasks from Adroit~\cite{rajeswaran2017learning} are evaluated. 
To learn the motor-control policy for each task, we train the behavior cloning networks with 3 different seeds, using the top-view inputs and the largest demo dataset size used in~\cite{nair2022r3m}. 
We report the average success rate of over 3 seeds for each task. 
All other configurations (e.g., network designs, training steps, and evaluation interval) are the same as those in R3M~\cite{nair2022r3m}.

For the multi-task setting, we evaluate our adapted pre-trained models on 18 multi-tasks from RLBench~\cite{james2020rlbench}. Existing pre-training works~\cite{nair2022r3m, xiao2022masked, huo2023human, dasari2023unbiased, jing2023exploring} for robotic manipulation only evaluate models under single-task setting, where the data scale, task diversity, and task complexity are limited.  
By additionally evaluating our method on the RLBench, which is challenging for its multi-task setting with significant task diversity and difficulty, the generalizability of our adapted models can be more comprehensively demonstrated.
Instead of using 8 self-attention layers like RVT~\cite{goyal2023rvt}, we only use one single self-attention layer to fuse the frozen image features and language features, and train the network for five epochs and test the network three times. Following RVT~\cite{goyal2023rvt}, we use a total of 1800 demonstrations for training and report the average success rate and standard deviation for testing.
All other experimental configurations are the same as RVT~\cite{goyal2023rvt}. 
Please see details of our downstream policy designs on RLBench in 
the appendix.

\noindent\textit{\textbf{(ii) Real-world tasks.}} 
For the real-world environment, we use an xArm7 robot arm, an Inspire gripper, and a third-view Orbbec Femto Bolt camera. Five tasks, i.e., \textit{put fruit in plate}, \textit{stack cups}, \textit{put tennis in mug}, \textit{hang mug}, and \textit{put item in box} are designed. Figure~\ref{fig:real_world} illustrates our experimental setups. For each task, we collect 40 demonstrations for training. During evaluation, we test 20 episodes for each task and report the average success rate.
For simplicity, we train the real-world policy under a single task setting, where the pre-trained models or our adapted models are used as visual feature extractors, with an ACT~\cite{zhao2023learning} framework to predict the following key-actions (likes RVT~\cite{goyal2023rvt}) of the end-effector.
For more details about the setups, data collection, and model designs, please refer to the appendix.

\subsection{Does the HR-Align method enable better generalization across different tasks?}
To demonstrate the effectiveness of our adaptation method on downstream tasks, we adapt two human-data pre-trained models that are widely used in the community, namely, \SecModel~\cite{dasari2023unbiased} and R3M~\cite{nair2022r3m}. 
We denote the pre-trained models after adaptation as \AdaptSecModel~and R3M-Align, respectively.

\noindent\textbf{Simulated single-task setting.} 
As shown in Table~\ref{tab:adroit}, we evaluate our adapted models on two tasks in the Adroit environment. 
For the \SecModel~model, our adapted version, namely the \AdaptSecModel~model, achieves improvements on both tasks, with an average success rate increased by $2.0\%$ compared to the unadapted \SecModel~model.
For the R3M model that was pre-trained on different human data with different pre-trained methodologies, our R3M-Align model also improves the success rate of both tasks, especially for the \textit{relocate} task (one of the most difficult tasks in Adroit). On average, our R3M-Align model has a significant $7.3\%$ improvement in success rate over the unadapted R3M model.

\vspace{-0.1cm}
\begin{table}[!h]
    \centering
    \setlength{\abovecaptionskip}{0.1cm}
    \resizebox{0.8\linewidth}{!}{ 
    \begin{tabular}{ccc>{\columncolor{successColor}}c}
    \specialrule{0.9pt}{0pt}{0pt}
    \multicolumn{1}{c|}{Models} & \textit{pen} & \textit{relocate} & Averaged \\
    \specialrule{0.9pt}{0pt}{0pt}
    \addlinespace[0.5ex]
    \specialrule{0.5pt}{0pt}{0pt}
    \multicolumn{1}{c|}{\SecModel} & 74.7 & 51.3 & 63.0 \\
    \multicolumn{1}{c|}{\AdaptSecModel} & \mybold{76.7} & \mybold{53.3} & \textbf{65.0 \myPink{(+2.0)}} \\
    \specialrule{0.5pt}{0pt}{0pt}
    \addlinespace[0.5ex]
    \specialrule{0.5pt}{0pt}{0pt}
    \multicolumn{1}{c|}{R3M} & 78.0 & 70.0 & 74.0 \\
    \multicolumn{1}{c|}{R3M-Align} & \mybold{81.3} & \mybold{81.3} & \textbf{81.3 \myPink{(+7.3)}} \\
    \specialrule{0.5pt}{0pt}{0pt}
    \end{tabular}
    }
\caption{\textbf{Success rate of tasks in Adroit}. Two pre-trained models, i.e., D4R and R3M, are adapted by our method. The adapted D4R-Align and R3M-Align models achieve better performance on the tasks compared to their unadapted counterparts.}
\label{tab:adroit}
\end{table}

\begin{table*}[!th]
\centering
\resizebox{1.0\linewidth}{!}{
\begin{tabular}{cccccccccccccccccccccc}
\specialrule{0.9pt}{0pt}{0pt}
\multicolumn{1}{c|}{\makecell[c]{Models}} 
                            &\begin{tabular}[c]{@{}c@{}}\texttt{put in} \\\texttt{drawer}\end{tabular}    
                            &\begin{tabular}[c]{@{}c@{}}\texttt{drag} \\\texttt{stick}\end{tabular} 
                            &\begin{tabular}[c]{@{}c@{}}\texttt{turn} \\\texttt{tap}\end{tabular}   
                            &\begin{tabular}[c]{@{}c@{}}\texttt{slide} \\\texttt{block}\end{tabular} 
                            &\begin{tabular}[c]{@{}c@{}}\texttt{open} \\\texttt{drawer}\end{tabular}    
                            &\begin{tabular}[c]{@{}c@{}}\texttt{put in}\\\texttt{cupboard}\end{tabular} 
                            &\begin{tabular}[c]{@{}c@{}}\texttt{sort}\\\texttt{shape}\end{tabular} 
                            &\begin{tabular}[c]{@{}c@{}}\texttt{put in}\\\texttt{safe}\end{tabular} 
                            &\begin{tabular}[c]{@{}c@{}}\texttt{push}\\\texttt{buttons}\end{tabular} 
                            &\begin{tabular}[c]{@{}c@{}}\texttt{close}\\\texttt{jar}\end{tabular}
                            \\
\specialrule{0.9pt}{0pt}{0pt}

\addlinespace[0.5ex]
\specialrule{0.5pt}{0pt}{0pt}
\multicolumn{1}{c|}{\SecModel} & $25.3_{\pm 6.1}$ & $100.0_{\pm 0.0}$ & $97.3_{\pm 4.6}$ & $57.3_{\pm 12.9}$ & $84.0_{\pm 10.6}$ & $46.7_{\pm 6.1}$ & $24.0_{\pm 4.0}$ & $88.0_{\pm 0.0}$ & $93.3_{\pm 2.3}$ & $61.3_{\pm 6.1}$ \\
\multicolumn{1}{c|}{\AdaptSecModel} & \mybold{$32.0_{\pm 4.0}$} & $97.3_{\pm 2.3}$ & $96.0_{\pm 0.0}$ & \mybold{$70.7_{\pm 2.3}$} & $81.3_{\pm 2.3}$ & \mybold{$46.7_{\pm 4.6}$} & $22.7_{\pm 2.3}$ & \mybold{$97.3_{\pm 2.3}$} & \mybold{$100.0_{\pm 0.0}$} & \mybold{$72.0_{\pm 6.9}$} \\
\specialrule{0.5pt}{0pt}{0pt}

\addlinespace[0.5ex]
\specialrule{0.5pt}{0pt}{0pt}
\multicolumn{1}{c|}{R3M} & $14.7_{\pm 4.6}$ & $96.0_{\pm 4.0}$ & $72.0_{\pm 4.0}$ & $60.0_{\pm 4.0}$ & $80.0_{\pm 4.0}$ & $24.0_{\pm 0.0}$ & $20.0_{\pm 4.0}$ & $90.7_{\pm 2.3}$ & $100.0_{\pm 0.0}$ & $74.7_{\pm 2.3}$  \\
\multicolumn{1}{c|}{R3M-Align} & \mybold{$70.7_{\pm 6.1}$} & \mybold{$97.3_{\pm 2.3}$} & \mybold{$80.0_{\pm 4.0}$} & \mybold{$80.0_{\pm 4.0}$} & \mybold{$80.0_{\pm 8.0}$} & \mybold{$33.3_{\pm 9.2}$} & $13.3_{\pm 2.3}$ & $78.7_{\pm 4.6}$ & \mybold{$100.0_{\pm 0.0}$} & \mybold{$80.0_{\pm 4.0}$} \\
\specialrule{0.5pt}{0pt}{0pt}

\\[1pt]  
\specialrule{0.9pt}{0pt}{0pt}
\multicolumn{1}{c|}{\makecell[c]{Models}} 
                            &\begin{tabular}[c]{@{}c@{}}\texttt{stack} \\\texttt{blocks}\end{tabular}
                            &\begin{tabular}[c]{@{}c@{}}\texttt{place}\\\texttt{cups}\end{tabular}
                            &\begin{tabular}[c]{@{}c@{}}\texttt{place}\\\texttt{wine}\end{tabular} 
                            &\begin{tabular}[c]{@{}c@{}}\texttt{screw}\\\texttt{bulb}\end{tabular} 
                            &\begin{tabular}[c]{@{}c@{}}\texttt{sweep to} \\\texttt{dustpan}\end{tabular}   
                            &\begin{tabular}[c]{@{}c@{}}\texttt{insert}\\\texttt{peg}\end{tabular} 
                            &\begin{tabular}[c]{@{}c@{}}\texttt{meat off} \\\texttt{grill}\end{tabular} 
                            &\begin{tabular}[c]{@{}c@{}}\texttt{stack}\\\texttt{cups}\end{tabular} 
                            &\multicolumn{2}{p{2.5cm}}{\cellcolor{successColor} \makecell[c]{Averaged \\ Success Rate}} 
                            \\
\specialrule{0.9pt}{0pt}{0pt}

\addlinespace[0.5ex]
\specialrule{0.5pt}{0pt}{0pt}
\multicolumn{1}{c|}{\SecModel} & $6.7_{\pm 2.3}$ & $0.0_{\pm 0.0}$ & $81.3_{\pm 6.1}$ & $32.0_{\pm 4.0}$ & $68.0_{\pm 6.9}$ & $32.0_{\pm 10.6}$ & $97.3_{\pm 2.3}$ & $0.0_{\pm 0.0}$ & \multicolumn{2}{p{2.5cm}}{\cellcolor{successColor}\makecell[c]{$55.3$}} \\
\multicolumn{1}{c|}{\AdaptSecModel} & \mybold{$16.0_{\pm 4.0}$} & $0.0_{\pm 0.0}$ & \mybold{$84.0_{\pm 4.0}$} & \mybold{$57.3_{\pm 8.3}$} & \mybold{$86.7_{\pm 4.6}$} & $24.0_{\pm 6.9}$ & $92.0_{\pm 4.0}$ & \mybold{$1.3_{\pm 2.3}$} & \multicolumn{2}{p{2.5cm}}{\cellcolor{successColor}\makecell[c]{\textbf{$59.9~\myPink{(+4.6)}$}}} \\

\specialrule{0.5pt}{0pt}{0pt}

\addlinespace[0.5ex]
\specialrule{0.5pt}{0pt}{0pt}
\multicolumn{1}{c|}{R3M} & $16.0_{\pm 4.0}$ & $0.0_{\pm 0.0}$ & $77.3_{\pm 4.6}$ & $30.7_{\pm 6.1}$ & $36.0_{\pm 4.0}$ & $17.3_{\pm 2.3}$ & $96.0_{\pm 0.0}$ & $0.0_{\pm 0.0}$ & \multicolumn{2}{p{2.5cm}}{\cellcolor{successColor}\makecell[c]{$50.3$}}\\
\multicolumn{1}{c|}{R3M-Align} & $13.3_{\pm 2.3}$ & $0.0_{\pm 0.0}$ & \mybold{$88.0_{\pm 4.0}$} & \mybold{$41.3_{\pm 4.6}$} & \mybold{$98.7_{\pm 2.3}$} & $16.0_{\pm 6.9}$ & $93.3_{\pm 2.3}$ & \mybold{$1.3_{\pm 2.3}$} & \multicolumn{2}{p{2.5cm}}{\cellcolor{successColor}\makecell[c]{\textbf{$59.2~\myPink{(+8.9)}$}}}\\
\specialrule{0.5pt}{0pt}{0pt}

\end{tabular}
}
\setlength{\abovecaptionskip}{0cm}
\caption{\textbf{Success rate of 18 tasks in RLBench}. Compared to the unadapted D4R and R3M models, our adapted D4R-Align and R3M-Align models significantly improve the success rate of most tasks. The average improvement of each model on all 18 tasks is also significant.}
\label{tab:rlbench}
\vspace{-0.3cm}
\end{table*}

\noindent\textbf{Simulated multi-task setting.}
Unlike previous works that evaluate models solely under the single-task setting, we conduct a more comprehensive evaluation of our method by evaluating the adapted pre-trained models on 18 language-conditioned tasks from the RLBench benchmark.
This multi-task setting is challenging, as it requires learning a single policy to complete various instructions in all $18$ manipulation tasks.
For each task, we report the average success rate and standard deviation over three tests.
The results are shown in the Table~\ref{tab:rlbench}. Compared to the unadapted \SecModel~model, our adapted \AdaptSecModel~model achieves significantly superior performance on $10$ tasks, with $4.6\%$ improvement in the averaged success rate of all $18$ tasks.
Similarly, our adapted R3M-Align model also significantly improves the success rate of $10$ tasks compared to the unadapted R3M model. More importantly, on this challenging benchmark, our adapted R3M-Align model achieves $8.9\%$ improvement in the averaged success rate of all $18$ tasks, demonstrating the necessity of mitigating the human-robot domain discrepancy in motor-control pre-training.

\noindent\textbf{Real-world task setting.} 
In the bottom right of Figure~\ref{fig:real_world}, we present the performance of the pre-trained D4R and R3M models, as well as our adapted D4R-Align and R3M-Align models, across the five tasks. 
For each task, the results indicate that our adapted models achieve significant improvements over their unadapted versions. 
On average, across all five tasks, the adapted D4R-Align model shows a $13\%$ improvement in success rate compared to the pre-trained D4R model, while the adapted R3M-Align model similarly demonstrates an $11\%$ improvement over the pre-trained R3M model. 
These results underscore the effectiveness of our adaptation method in improving existing visual pre-trained models in real-world scenarios.

\subsection{How effective is the design of human-robot cross-domain alignment?}
Our HR-Align method adapts existing human-data pre-trained models in an efficient way to better generalize the new representation across different downstream tasks. 
To mitigate the human-robot discrepancy, the semantic alignment design upon the paired human-robot video dataset is the key to efficiently adapting the pre-trained models. 
To validate this, we demonstrate two strong baselines by fully tuning the existing pre-trained model (i.e., R3M) on all robot videos in the paired human-robot dataset, with different fine-tuning objectives. 
The first one, since the R3M model is pre-trained on Ego4D videos, to mitigate the human-robot domain gap, we continue to finetune the R3M model on the used robot videos using its original pre-training method. We call this newly adapted model R3M-PreT.
For the second one, as the actions performed in the used robot videos are well-defined, this indicates the videos contain rich manipulation-oriented semantics. To this end, we finetune the pre-trained R3M model by classifying each robot video into its action category, making the learned feature representation more effective in capturing visual dynamics. We call this adapted model R3M-ClS.

\vspace{-0.3cm}
\begin{table}[!h]
\begin{center}
\resizebox{1.0\linewidth}{!}
{
\begin{tabular}{cccc>{\columncolor{successColor}}c}
\specialrule{0.9pt}{0pt}{0pt}
\multicolumn{1}{c|}{Models} & learned params. & \textit{pen} & \textit{relocate} & Averaged \\
\specialrule{0.9pt}{0pt}{0pt}

\addlinespace[0.5ex]
\specialrule{0.5pt}{0pt}{0pt}
\multicolumn{1}{c|}{R3M} & 0M (frozen:25M) & 78.0 & 70.0 & 74.0 \\
\multicolumn{1}{c|}{R3M-PreT} & 25M & 78.0 & 77.3 & 77.7 (+3.7) \\
\multicolumn{1}{c|}{R3M-ClS} & 25M & 79.3 & 75.3 & 77.3 (+3.3)\\
\multicolumn{1}{c|}{R3M-Align} & 1.6M & \textbf{81.3} & \textbf{81.3} & \textbf{81.3 \myPink{(+7.3)}} \\
\specialrule{0.5pt}{0pt}{0pt}

\specialrule{0.5pt}{0pt}{0pt}
\end{tabular}
}
\end{center}
\setlength{\abovecaptionskip}{-0.2cm}
\setlength{\belowcaptionskip}{-0.3cm}
\caption{Success rates of two tasks in \textbf{Adroit}. The R3M-PreT and R3M-ClS are two strong baselines by fully fine-tuning on the used robot videos. Our R3M-Align model with human-robot semantic alignment achieves more superior performance by tuning much fewer parameters.}
\label{ab:hr_align}
\end{table}
\vspace{-0.1cm}

As shown in Table~\ref{ab:hr_align}, on the two tasks in Adroit, both the R3M-PreT and R3M-ClS models improve the manipulation success rate, demonstrating that fine-tuning pre-trained models on the robot data is effective in mitigating the human-robot domain discrepancy in pre-training.
Nevertheless, all parameters of the visual backbones in these two models are required to be trained, making the adaptation process inefficient.
Compared to these two models, with the proposed HR-Align method, our R3M-Align model achieves more significant improvements on the two tasks, by tuning only $6.4\%$ of the parameters.
This demonstrates that our human-robot semantic alignment design is more effective in addressing the domain discrepancy problem in pre-training. And more importantly, it provides a simple yet effective way to adapt existing pre-trains in an efficient manner.

\subsection{How much does each component contribute to the HR-Align method?}
Driven by the idea of human-robot semantic alignment, another contribution of this work is leveraging the parameter-efficient design (i.e., adapter) to adapt the representation of pre-trained models.
Following existing practice~\cite{sharma2022lossless, yang2023aim, pan2022st}, for a visual backbone consisting of multiple blocks (e.g., the residual blocks in resnet~\cite{he2016deep}), we selectively add the adapter modules at three positions in the backbone network. 
Taking the pre-trained R3M model as an example.
We can insert our adapter modules before the first block, in the middle of all blocks, and after the last block. We call these three adapted models R3M-Align-E, R3M-Align-M, and R3M-Align-L, respectively.
Similarly, we also add the adapter modules at all three locations of the R3M model, which we denote as R3M-Align-E.M.L.

In Table~\ref{ab:adapter_pos_and_lang}, we evaluate the performance of these models on two tasks in the Adroit environment.
Compared to the R3M model which has 25 million parameters in its backbone, all four adapted models can significantly improve the success rate of downstream tasks by using very few learnable parameters. 
For instance, the R3M-Align-L model achieves maximum improvement by adding only $6.4\%$ of parameters.
Interestingly, compared to the other three models that insert the adapter module at a single position, the R3M-Align-E.M.L model does not perform the best while using more learnable parameters. This phenomenon indicates that it is difficult to harmoniously adapt representations at multiple positions of the pre-trained model for aligning semantics between human and robot data.

\vspace{-0.2cm}

\begin{table}[!th]
\begin{center}
\resizebox{1.0\linewidth}{!}
{
\begin{tabular}{cccc>{\columncolor{successColor}}c}
\specialrule{0.9pt}{0pt}{0pt}
\multicolumn{1}{c|}{Models} & learned params. & \textit{pen} & \textit{relocate} & Averaged \\
\specialrule{0.9pt}{0pt}{0pt}

\addlinespace[0.5ex]
\specialrule{0.5pt}{0pt}{0pt}
\multicolumn{1}{c|}{R3M} & 0M (frozen:25M) & 78.0 & 70.0 & 74.0 \\
\specialrule{0.5pt}{0pt}{0pt}

\addlinespace[0.5ex]
\specialrule{0.5pt}{0pt}{0pt}
\multicolumn{1}{c|}{R3M-Align-E} & 0.1M & 75.3 & \textbf{84.0} & 79.6 (+5.6) \\
\multicolumn{1}{c|}{R3M-Align-M} & 3.5M & 80.7 & 82.0 & 81.3 (+7.3) \\
\multicolumn{1}{c|}{R3M-Align-L} & 1.6M & \textbf{81.3} & 81.3 & \textbf{81.3 \myPink{(+7.3)}} \\
\multicolumn{1}{c|}{R3M-Align-E.M.L} & 5.2M & 77.3 & 83.3 & 80.3 (+6.3) \\
\specialrule{0.5pt}{0pt}{0pt}
\end{tabular}
}
\end{center}
\setlength{\abovecaptionskip}{-0.2cm}
\caption{Success rates of two tasks in \textbf{Adroit}. Ablation studies on inserting adapter modules at different positions in the R3M model.}
\label{ab:adapter_pos_and_lang}
\end{table}

\begin{figure}[h]
  \centering
   \includegraphics[width=0.9\linewidth]{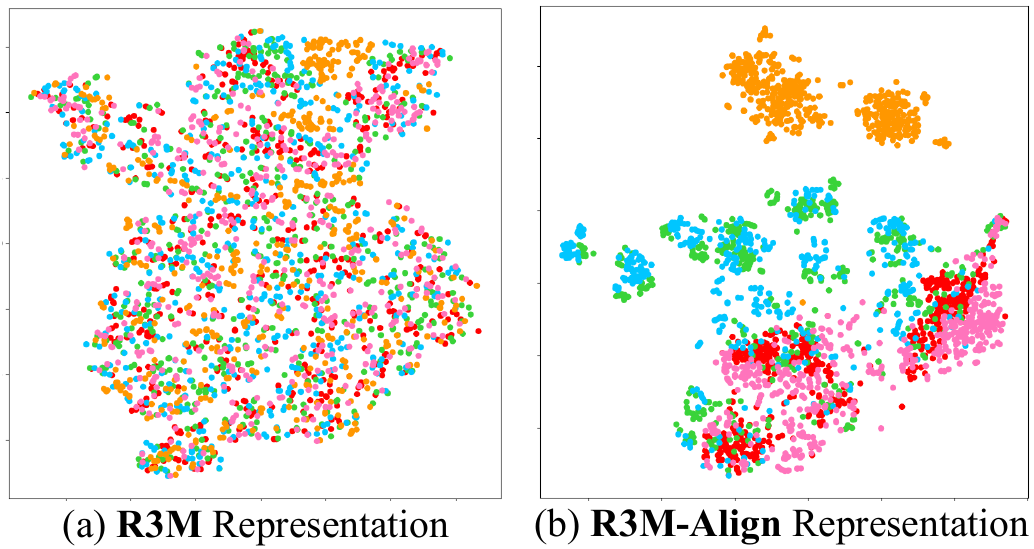}
   \caption{The t-SNE visualizations of the RLBench's feature distributions of R3M and R3M-Align models. Each color denotes a task, and the points denote different samples.}
\label{fig:feat_vis}
\end{figure}

\subsection{Is the adapted representation more effective on downstream tasks?}
To investigate the effectiveness of our proposed adaptation method, we employ t-SNE~\cite{van2008visualizing} to depict the downstream feature distributions of the existing pre-trained R3M model and our adapted R3M-Align model, respectively.
We randomly select 5 tasks from the RLBench dataset, where 500 samples are randomly selected for each task. 
The distributions of feature representation of the R3M model and R3M-Align model are shown in Figure~\ref{fig:feat_vis} (a) and Figure~\ref{fig:feat_vis} (b), respectively. In the figure, different colors represent different tasks, and each point represents a sample.
In the visualization, we find that the feature distribution of the existing R3M model is significantly more dispersed than that of our adapted R3M-Align model.
This phenomenon demonstrates that our adaptation method can effectively learn more discriminative representation for downstream tasks by addressing the human-robot domain discrepancy in pre-training.


\section{Conclusions and Limitations}
\label{sec:conclusions}
Learning generalizable representations from large-scale human data for robotic manipulation shows great potential.
However, human-robot domain discrepancy remains a challenge that existing pre-training paradigms cannot effectively address. 
This work makes a preliminary attempt to tackle this challenge by introducing a new adaptation paradigm, which leverages paired human-robot data and proposes an efficient semantic alignment method. 
Through this approach, human-data pre-trained models can be explicitly and efficiently adapted to the robot domain without the need to be tailored for each downstream robotic environment.
Experiments on 20 simulated manipulation tasks and five real-world manipulation tasks with different pre-trained models, demonstrate the effectiveness of our proposed method.

To motivate further research, we highlight several limitations of this work. 
First, exploring ways to measure domain discrepancies between human-data pre-training and downstream robotic tasks could guide the adaptation process and potentially improve outcomes. 
Second, this work focuses on adapting pre-trained models without revisiting the pre-training stage. Integrating paired human-robot data with out-of-domain human data for pre-training could lead to more advanced models and adaptation methods. 
Finally, further research is needed to understand how the scale and diversity of both pre-training data and paired human-robot data impact domain discrepancies in pre-training for robotic manipulation.

\section*{Acknowledgement}
\noindent This work was supported by the National Natural Science Foundation of China (No. 62306257), and the Guangzhou Municipal Science and Technology Project (No. 2024A03J0619 \& 2024A04J4390). This work was also supported by the Guangzhou-HKUST(GZ) Joint Funding Program (Grant No.2023A03J0008) and Education Bureau of Guangzhou Municipality.

\newpage

\appendix

\renewcommand\thefigure{{S\arabic{figure}}}
\renewcommand\thetable{{S\arabic{table}}}
\renewcommand\thesection{{S\arabic{section}}}

\setcounter{figure}{0} 
\setcounter{table}{0} 
\setcounter{section}{0} 

{\Large\textbf{Appendix}}

\section{Downstream Policy Learning on RLBench}
\label{appendix:policy_rlbench}
\begin{figure*}[!h]
  \centering
   \includegraphics[width=0.95\linewidth]{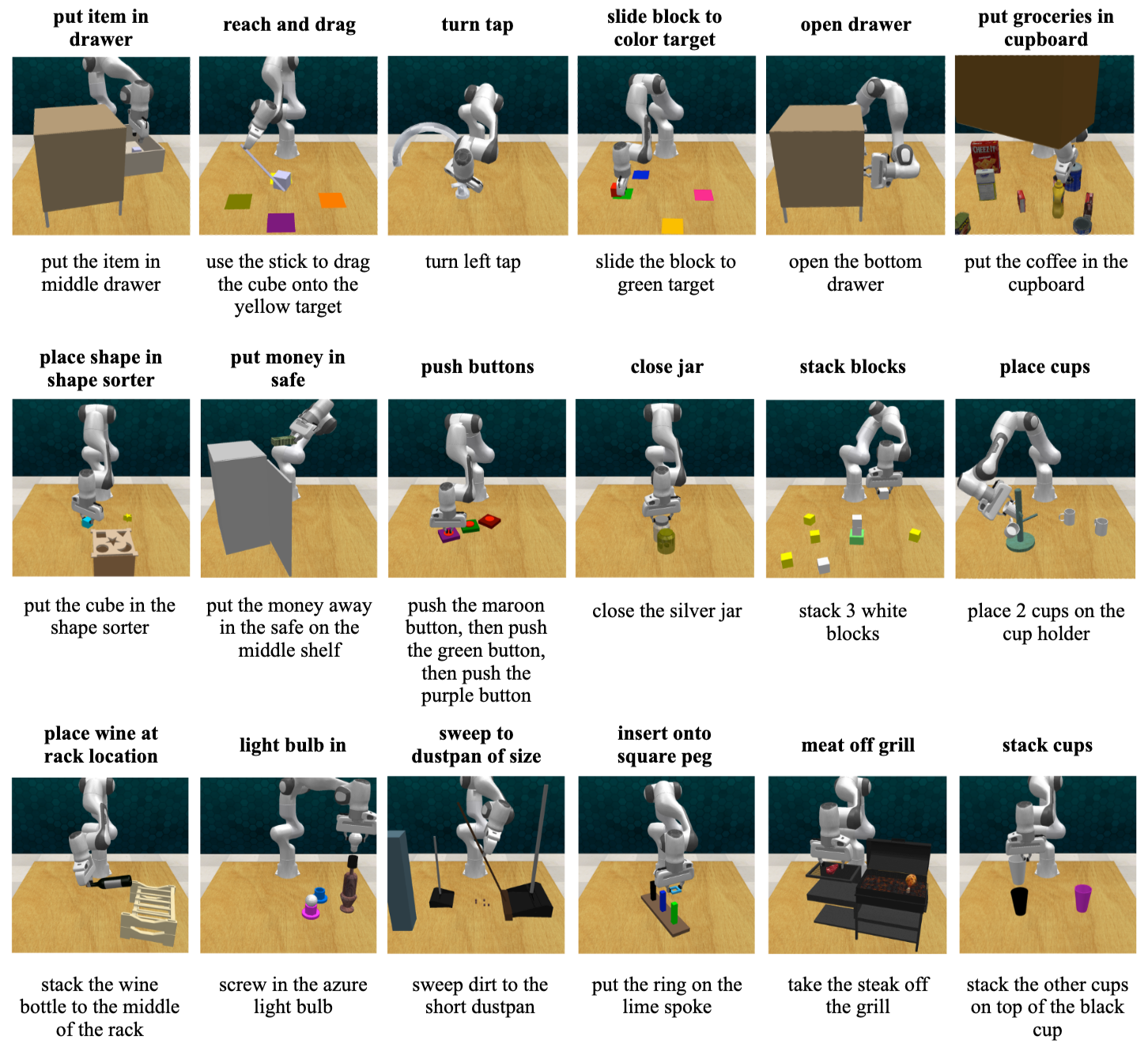}
   \caption{Examples of the 18 RLBench tasks (front view) with corresponding human instructions (sourced from~\cite{ma2024contrastive}).}
\label{fig:rlbench}
\end{figure*}

Existing visual pre-training works often assess the learned representation in downstream environments using a single-task setting.
However, our work diverges from this standard by also evaluating the adapted pre-trained models on the large-scale RLBench benchmark~\cite{james2020rlbench}, where a single language-conditioned policy is learned to complete various tasks. Figure~\ref{fig:rlbench} shows examples of the 18 tasks and the corresponding human instructions.

On RLBench, existing works~\cite{chen2023polarnet, goyal2023rvt, he2024large, james2022q, shridhar2023perceiver} usually develop sophisticated models to model the semantics of the robot's multi-view observations and their correlations with the language commands.
For example, RVT~\cite{goyal2023rvt} utilizes four attention layers as the visual encoder to model intra-image relations and four more attention layers to model image-language correlations.
In this work, we adopt the same design as RVT, but replace the visual encoder (i.e., four intra-image attention layers) with either an existing pre-trained model or our adapted one.
Additionally, we employ just one attention layer to fuse the extracted image features and language features.
Since RVT predicts the end-effector's actions based on the features without down-sampling the spatial dimension, we discard all spatial down-sampling operations (e.g., max-pooling) in both the pre-trained models and our adapted models.
Please note that to validate the effectiveness of our adaptation method, we freeze the visual representation of the pre-trained model or our adapted model while learning the downstream policy on RLBench.

\begin{figure*}[!t]
  \centering
   \includegraphics[width=1.0\linewidth]{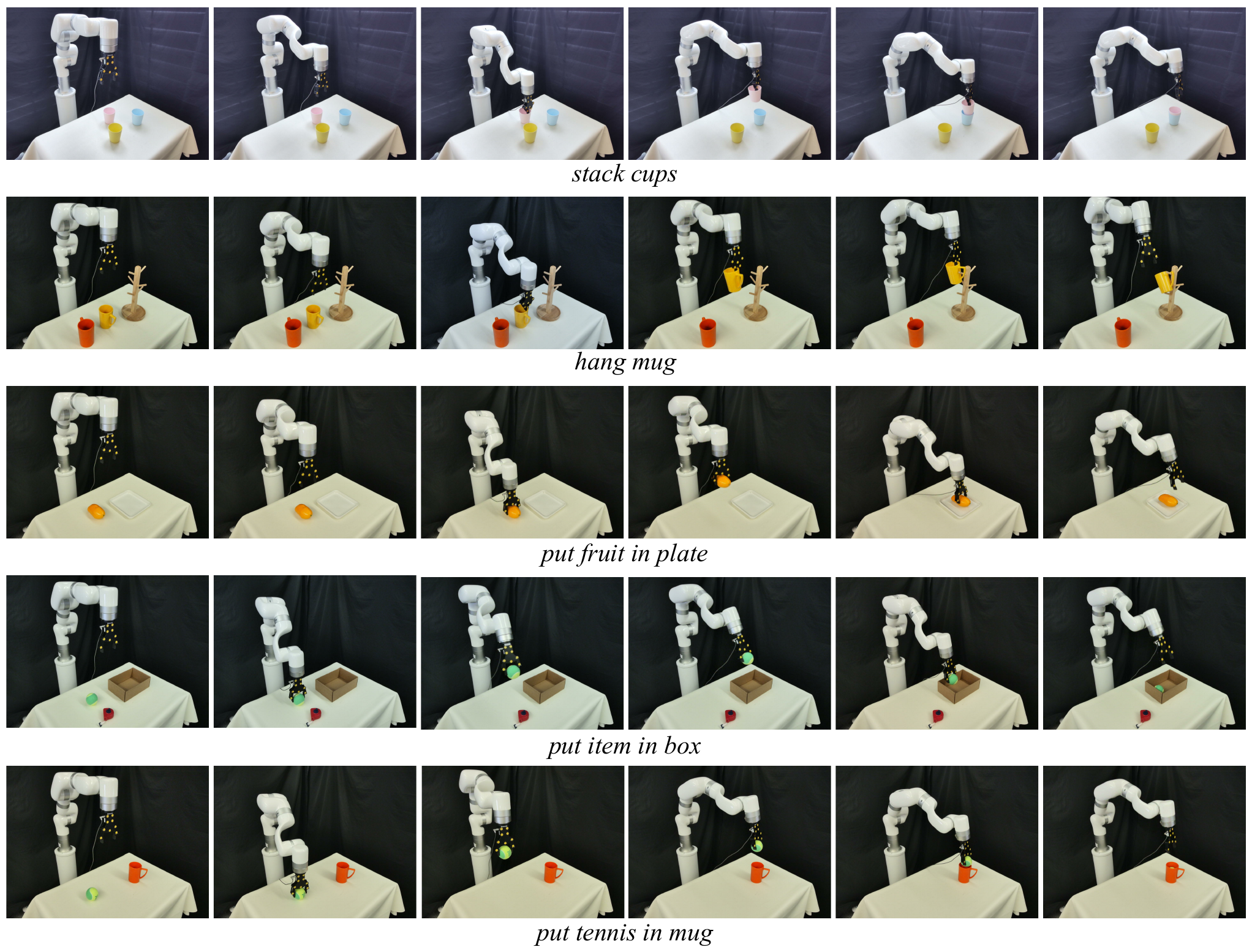}
   \caption{Examples of the five real-world tasks are shown, with each row presenting an instance of the corresponding task. For each demonstration, we provide visual observations at six different timestamps.}
\label{fig:real_tasks}
\end{figure*}

\section{Real-world Experiments}
\label{appendix:real_world}

\noindent\textbf{Setups.}
For the real-world manipulation experiments, we use a 7-DoF xArm robot arm equipped with an Inspire gripper. 
Visual observations are captured by an Orbbec Femto Bolt (RGB-D) camera positioned in front of and to the upper right of the robot arm. 
The positions of the robot arm, working area, and camera remain fixed during data collection and policy testing. 
Additionally, we use a DJI Osmo Action 4 camera to record videos of the policy testing.

\noindent\textbf{Data Collection.}
We design five different real-world tasks, namely, \textit{put fruit in plate}, \textit{stack cups}, \textit{put tennis in mug}, \textit{hang mug}, and \textit{put item in box}.
For each task, we collect 40 human teleoperation demonstrations for training.
Figure~\ref{fig:real_tasks} demonstrates some examples of the collected tasks.
For each demonstration, we manually move the robot arm and change the states of the gripper (i.e., open or close) to complete the target task.
We record these operations and replay them to record the demonstration.
We simultaneously record the robot arm end-effector state (i.e., positions in the x-axis, y-axis, and z-axis, and rotations in roll, pitch, and yaw), gripper state, and Orbbec camera RGB stream with an image size of 1280$\times$960.

\noindent\textbf{Model Designs.}
We train our manipulation policy network under a single-task setting, utilizing the ACT~\cite{zhao2023learning} framework for policy learning\footnotemark. In addition, following RVT~\cite{goyal2023rvt} which predicts the next key-action, we predict the following key-actions of the end-effector. 
The visual backbone of the network is replaced with either the pre-trained models or our adapted models. 
During both training and testing, the RGB images are resized to $320 \times 240$.

\footnotetext{We follow the implementation of https://github.com/Shaka-Labs/ACT.}

\begin{table}[!h]
\begin{center}
\resizebox{1.0\linewidth}{!}
{
\begin{tabular}{cccc>{\columncolor{successColor}}c}
\specialrule{0.9pt}{0pt}{0pt}
\multicolumn{1}{c|}{Models} & learned params. & \textit{pen} & \textit{relocate} & Averaged \\
\specialrule{0.9pt}{0pt}{0pt}

\addlinespace[0.5ex]
\specialrule{0.5pt}{0pt}{0pt}
\multicolumn{1}{c|}{R3M} & 0M (frozen:25M) & 78.0 & 70.0 & 74.0 \\
\specialrule{0.5pt}{0pt}{0pt}

\addlinespace[0.5ex]
\specialrule{0.5pt}{0pt}{0pt}
\multicolumn{1}{c|}{R3M-Align-L} & 1.6M & \textbf{81.3} & \textbf{81.3} & \textbf{81.3 \myPink{(+7.3)}} \\
\multicolumn{1}{c|}{R3M-Align-L \textit{w/o lang.}} & 1.6M & 79.3 & 80.7 & 80.0 (+6.0) \\

\specialrule{0.5pt}{0pt}{0pt}
\end{tabular}
}
\end{center}
\setlength{\abovecaptionskip}{0cm}
\caption{Success rates of two tasks in \textbf{Adroit}. Removing the language-guided feature enhancement will degrade the model's performance.}
\label{ab:lang_supp}
\end{table}

\section{Additional Ablations}
In this work, our adaptation method uses task description features as queries to better capture task-aware semantics from video features.
As shown at the bottom of Table~\ref{ab:lang_supp}, by removing this operation, the adapted model, i.e., R3M-Align-L \textit{w/o lang.}, will result in performance degradation. This demonstrates that the language-guided feature enhancement is effective in promoting human-robot semantic alignment.

In Table 3, we only used the robot data from the RH20T subset we used to train R3M-PreT and R3M-ClS. For fair comparisons, we train both R3M-PreT and R3M-ClS using human and robot videos (i.e., the same amount of training data of HR-Align).
Table~\ref{tab:fair_comparisons_data} shows that our R3M-Align still outperforms R3M-PreT and R3M-ClS trained by full-data. 
In addition, the R3M-PreT and R3M-ClS in Table 3 are full-parameter fine-tuned, while our HR-Align is fine-tuned with parameter-efficient Adapter.
To ensure a fair comparison, we instantiate R3M-PreT and R3M-ClS by inserting an adapter into frozen R3M (i.e., the same amount of learnable parameters as our R3M-Align), and train them using both human and robot data (i.e., the same amount of training data of HR-Align), denoted as $\text{R3M-PreT}_{A}$ and $\text{R3M-ClS}_{A}$.
Table~\ref{tab:fair_comparisons_param} shows that our R3M-Align still performs better. The above shows the effectiveness of our HR-align method.

\begin{table}[!h]
\centering
\begin{tabular}
{c|c|c}
\hline
R3M-PreT & R3M-ClS & R3M-Align (\textbf{Ours}) \\
\hline
 78.1 & 77.5 & \textbf{81.3} \\
\hline
\end{tabular}
\caption{Comparisons between models when training with full data.}
\label{tab:fair_comparisons_data}
\end{table}

\begin{table}[!h]
\centering
\begin{tabular}
{c|c|c}
\hline
R3M-PreT$_A$ & R3M-ClS$_A$ & R3M-Align (\textbf{Ours}) \\
\hline
77.2 & 76.9 & \textbf{81.3}\\
\hline
\end{tabular}
\caption{Comparisons between models when training with Adapters.}
\label{tab:fair_comparisons_param}
\vspace{-0.3cm}
\end{table}

{
    \small
    \bibliographystyle{ieeenat_fullname}
    \bibliography{main}
}

\end{document}